\pdfoutput=1
\documentclass[11pt]{article}
\usepackage{graphicx} 
\usepackage{diagbox}
\usepackage{authblk}
\makeatletter
\renewcommand\AB@authnote[1]{\textsuperscript{#1}\hspace{5pt}}

\makeatother
\usepackage{hyperref}
\usepackage{algorithm}
\usepackage{algpseudocode}
\usepackage{notation-2}
\usepackage{arxiv-2}
\usepackage{amssymb}
\usepackage{mathrsfs}
\usepackage{float}
\usepackage{setspace}
\usepackage{lmodern}

\renewcommand{\tilde}[1]{\widetilde{#1}}

\renewcommand{\hat}[1]{\widehat{#1}}

\title{\normalfont Tight Sequential Confidence Intervals for Online Kernel Regression
}
\begin{centering}
     \author[1,2]{Haichen Hu\thanks{{Email: \texttt{huhc@mit.edu}}}}
      \author[2,3]{David Simchi-Levi\thanks{{Email: \texttt{dslevi@mit.edu}}}}
\end{centering}
\affil[1]{Center for Computational Science and Engineering, MIT}
\affil[2]{Department of Civil and Environmental Engineering, MIT}
\affil[3]{Institute for Data, Systems, and Society, MIT}
\begin{document}
\maketitle
\singlespacing
\begin{abstract}
We characterize, in the worst case over kernels, the sharp minimax width of online confidence intervals for elements of a reproducing kernel Hilbert space (RKHS). This resolves the open problem on tight online confidence intervals posed by \citet{vakili2021open}, in its original discrete-time form.
A learner observes \(y_i=f(x_i)+\epsilon_i\) at adaptively chosen points,
with \(\|f\|_{\mathcal H_k}\le B\) and \(R\)-sub-Gaussian noise, and reports
the interval \(|f(x)-\mu_n(x)|\le\rho_n(\delta)\sigma_n(x)\), where \(\mu_n\)
and \(\sigma_n\) are the posterior mean and standard deviation of a Gaussian
process model with regularization parameter \(\lambda\). Writing
\(\rho_n^*(\delta)\) for the minimax confidence width for a fixed kernel,
we show that, for the sum of a linear kernel and a periodic Sobolev kernel,
all sufficiently large \(n\), and all \(\delta\in(0,1/4]\),
\[
\rho_n^*(\delta)
\asymp
B+\frac{R}{\lambda}\sqrt{\gamma_n+\log\frac1\delta},
\]
where \(\gamma_n\) is the maximal information gain, which grows polynomially
in \(n\). Building on \citet{lattimore2023lower}, we construct an adaptive design on a fixed two-dimensional domain
for which, under Gaussian noise, the normalized error at a prespecified
point is \(\Omega(B+(R/\lambda)\sqrt{\gamma_n})\) with constant probability.
This rules out the conjectured polylogarithmic width in the worst case
over kernels and online designs.
\end{abstract}

\section{Introduction}
\label{sec:introduction}

Kernel methods extend linear models to rich nonlinear function classes, and a
large body of work studies online learning problems in which the unknown
objective lies in a reproducing kernel Hilbert space (RKHS). Two prominent
applications are kernelized bandits
\citep{srinivas2010gaussian,valko2013finite,chowdhury2017kernelized,scarlett2017lower,vakili2021information}
and kernel-based reinforcement learning
\citep{yang2020provably,vakili2023kernelized,vakili2024open,hu2026minimax}.

In kernelized bandits and kernel-based reinforcement learning, surrogate
Gaussian process (GP) models are often used to construct confidence
intervals.

We observe an unknown function \(f\in\mathcal H_k\) on a compact domain
\(\mathcal X\subset\mathbb R^d\) through
\begin{equation}
\label{eq:intro-model}
y_i=f(x_i)+\epsilon_i,
\ i=1,\ldots,n,
\ 
\|f\|_{\mathcal H_k}\le B,
\end{equation}
where \(k\) is a known kernel and the noise terms \(\epsilon_i\) are
i.i.d.\ \(R\)-sub-Gaussian. The design is \emph{online}, so \(x_i\) may
depend on the past observations \(\{(x_j,y_j)\}_{j<i}\).
Following \citet{vakili2021open}, let \(\mu_n\) and \(\sigma_n\) denote
the posterior mean and standard deviation of a zero-mean surrogate GP
with kernel \(k\) and regularization parameter \(\lambda>0\).
We seek the smallest width \(\rho_n(\delta)\) such that, for every
admissible \(f\), noise law, and online design, the following holds with
probability at least \(1-\delta\).
\begin{equation}
\label{eq:intro-ci}
|f(x)-\mu_n(x)|\le\rho_n(\delta)\,\sigma_n(x)
\ \text{ for all }x\in\mathcal X.
\end{equation}

For offline designs independent of the noise, standard confidence bounds
have only logarithmic dependence on \(n\) under suitable kernel regularity
conditions \citep{vakili2021optimal,vakili2021open}.
In the online setting, self-normalized concentration gives the width
\citep{abbasi2011improved,abbasi2013thesis,chowdhury2017kernelized,durand2018streaming}
\[
\rho_n(\delta)= 
B+\frac{R}{\lambda}\sqrt{2\log\frac1\delta+\gamma_n},
\]
where \(\gamma_n\) is the maximal information gain after \(n\) observations.
For kernels with polynomially decaying eigenvalues, including Mat\'ern
kernels, \(\gamma_n\) can grow polynomially in \(n\), making this width
substantially larger than its offline counterpart
\citep{vakili2021information,janz2026maximum}.

Over \(N\) rounds, the standard analysis of the Gaussian process upper
confidence bound (GP-UCB) algorithm gives regret
\(\tilde O(\rho_N(\delta)\sqrt{N\gamma_N})\), which becomes
\(\tilde O(\gamma_N\sqrt N)\) under the existing confidence width
\citep{srinivas2010gaussian,chowdhury2017kernelized}.
SupKernelUCB achieves \(\tilde O(\sqrt{N\gamma_N})\) using offline confidence
intervals \citep{valko2013finite}. This gap motivates asking whether the
\(\sqrt{\gamma_n}\) term in online confidence widths is necessary.

\citet{vakili2021open} asked whether the \(\sqrt{\gamma_n}\) term in online
confidence widths could be replaced by a term that is polylogarithmic in \(n\).
The discrete-time lower bound for linear regression of
\citet{khamaru2021near} does not distinguish these rates, since
\(\gamma_n\) grows only logarithmically in \(n\) for fixed dimension.
\citet{lattimore2023lower} showed that this inflation is unavoidable in a
continuous-time analogue, but explicitly mentioned that the corresponding
discrete-time result remained unproved despite supporting simulations.
Thus, the original question remained open for kernels with polynomially
growing information gain:

\begin{center}
\emph{
What is the minimax optimal growth rate of the width $\rho_n(\delta)$ with $n$? In
particular, can the existing width be reduced by a factor of order
$\sqrt{\gamma_n}$, up to logarithmic factors?
}
\end{center}

We resolve this question in its original discrete-time model, in the worst
case over kernels. For each smoothness scale \(s>1/2\), we construct the sum of a linear
kernel and a periodic Sobolev kernel of order \(s\) on the fixed domain
\(\mathcal X=[-1,1]\times[0,2\pi]\subset\mathbb R^2\).
For \(f_B(a,b)=-Ba\) and i.i.d.\ \(\mathcal N(0,R^2)\) noise, we construct,
for each \(n\ge n_0(s,\lambda)\), an online design such that, with
probability at least \(6/11\),
\[
\frac{|f_B(x_0)-\mu_n(x_0)|}{\sigma_n(x_0)}
\gtrsim
B+\frac{R}{\lambda}\Big(\frac{n}{\lambda^2}\Big)^{\frac{1}{4s}}
\asymp
B+\frac{R}{\lambda}\sqrt{\gamma_n},
\]
where \(x_0=(1,0)\) is fixed before observing any data.

Let \(\rho_n^*(\delta)\) denote the minimax confidence width of
\eqref{eq:intro-ci}, defined in Section~\ref{sec:setup}.
Combining this construction with the existing upper bound and a Gaussian
tail lower bound for a deterministic design yields, for all
\(n\ge n_0(s,\lambda)\) and \(\delta\in(0,1/4]\),
\[
\rho_n^*(\delta)
\asymp
B+\frac{R}{\lambda}\sqrt{\gamma_n+\log\frac1\delta},
\ 
\gamma_n\asymp\Big(\frac{n}{\lambda^2}\Big)^{\frac{1}{2s}},
\]
where all comparison constants depend only on \(s\).

Since our kernel has input dimension \(d=2\) and polynomially growing
information gain, the conjectured polylogarithmic width cannot hold for
every kernel. Consequently, the gap in the standard GP-UCB regret analysis
cannot be closed in general solely by tightening a confidence width that
is valid for every kernel and online design and depends on the kernel only
through \(\gamma_n\) and \(d\). Whether the \(\sqrt{\gamma_n}\) dependence
can be improved for Mat\'ern or squared exponential kernels remains open.

\paragraph{Notation:}
We write \([n]=\{1,\ldots,n\}\) and use \(|J|\) for the cardinality of a
finite set \(J\). For a vector \(v\), \(\|v\|_2\) denotes its Euclidean
norm. For a matrix \(A\), \(\|A\|_{\mathrm{op}}\), \(\|A\|_F\), and
\(\tr(A)\) denote its operator norm, Frobenius norm, and trace, respectively.
For symmetric matrices, \(A\succeq A'\) means that \(A-A'\) is positive
semidefinite, and \(A\succ0\) means that \(A\) is positive definite.
We use \(I_n\) for the \(n\times n\) identity matrix, \(\one_n\) for the
all-ones vector in \(\RR^n\), and \(\otimes\) for the Kronecker product.
For \(A\in\RR^{n\times n}\) and \(J\subseteq[n]\), \(A_J\) denotes the
principal submatrix with rows and columns indexed by \(J\).
The notation \(\mathcal N(\mu,\sigma^2)\) denotes the Gaussian distribution
with mean \(\mu\) and variance \(\sigma^2\), and \(\mathcal N(0,I_n)\)
denotes the standard Gaussian distribution on \(\RR^n\).
We use \(\floor{\cdot}\) and \(\ceil{\cdot}\) for the floor and ceiling
functions, and \(\mathrm{i}\) for the imaginary unit.
For nonnegative quantities \(u,v\), we write \(u\gtrsim v\) if
\(u\ge cv\) and \(u\asymp v\) if \(cv\le u\le Cv\), where \(c,C>0\)
depend at most on the smoothness parameter \(s\), unless stated otherwise.
The symbols \(O(\cdot)\) and \(\Omega(\cdot)\) have their usual asymptotic
meanings as \(n\to\infty\), and \(\tilde O(\cdot)\) hides polylogarithmic
factors.

\section{Literature Review}
\label{sec:related-work}

\paragraph{Kernelized bandits and Gaussian process optimization.}
\citet{srinivas2010gaussian} expressed GP-UCB regret through maximal
information gain, with subsequent RKHS and contextual extensions
\citep{chowdhury2017kernelized,krause2011contextual}.
Lower bounds were established for the squared exponential and Mat\'ern
kernels \citep{scarlett2017lower,cai2021lower}.
More recently, \citet{ma2026general} prove an
\(\Omega(\sqrt{N\gamma_N/\log N})\) expected regret lower bound for
nonconstant continuous kernels on compact domains under Gaussian noise.

For the Mat\'ern family, domain partitioning \citep{janz2020bandit} and
horizon-dependent regularization \citep{whitehouse2023sublinear} yield
sublinear regret. GP-UCB also admits improved guarantees in noise-free
RKHS models and in the Bayesian setting
\citep{iwazaki2025noise,iwazaki2025improved}, while regret lower bounds
are available for specific exploration and regularization schedules
\citep{wang2026suboptimality,xu2026algorithmic}.

Related questions arise in kernel-based reinforcement learning
\citep{yang2020provably,vakili2023kernelized,vakili2024open}.
For Sobolev RKHS reward and transition-density models,
\citet{hu2026minimax} characterize the minimax expected regret as a
function of the number of episodes.

\paragraph{Self-normalized concentration and online confidence intervals.}
Confidence intervals and high-probability bounds are fundamental tools in
probability and statistics. These include time-uniform confidence
sequences for sequential data
\citep{howard2021time,ryu2024gambling} and uniform or pointwise bounds for
stochastic processes
\citep{talagrand2005generic,dirksen2015tail,xu2026pointwise,hu2026simultaneous}.
For online regression, self-normalized martingale inequalities
\citep{delapena2004self,delapena2009self} underpin confidence ellipsoids for
linear bandits and their Hilbert-space extensions
\citep{abbasi2011improved,abbasi2013thesis}.
They yield online confidence bounds with a \(\sqrt{\gamma_n}\) dependence
for kernel regression
\citep{chowdhury2017kernelized,durand2018streaming,whitehouse2023sublinear}.
Recent work develops tighter kernel confidence bounds
\citep{flynn2024tighter}, dimension-free Bernstein-type inequalities
\citep{akhavan2025dimension}, bounds without regularization penalties
under additional covariate assumptions \citep{chen2026selfnormalized},
and guarantees under kernel misspecification \citep{maran2026sharper}.

\paragraph{Maximal information gain.}
The maximal information gain was introduced by \citet{srinivas2010gaussian},
with bounds for common kernels building on \citet{seeger2008information}.
\citet{vakili2021information} bound it through the eigenvalue decay of the
kernel, which gives \(\gamma_n=\tilde O(n^{d/(2\nu+d)})\) for the
Mat\'ern-\(\nu\) kernel, and \citet{janz2026maximum} determine its sharp
order, up to constant factors, for the Mat\'ern and squared exponential
kernels. More broadly, effective dimension and eigenvalue decay control
regret bounds in kernelized contextual bandits and online decision-making
with functional regression
\citep{valko2013finite,zenati2022efficient,neu2024adversarial,hu2025contextual}.
For our constructed kernel, we obtain matching information-gain bounds
using Hadamard's inequality for the upper bound and the spectrum of a
repeated equispaced design for the lower bound.

\paragraph{Alternative approaches to kernelized bandits.}
In kernelized bandits, SupKernelUCB, batched elimination, and domain
shrinking use staged or batched data collection to apply offline confidence
bounds \citep{valko2013finite,li2022gaussian,salgia2021domain}.
Related approaches combine experimental design with robust estimation
\citep{camilleri2021high}, or use response-independent variance
maximization to achieve near-optimal simple regret
\citep{vakili2021optimal}.
More recently, \citet{takeno2026near} use data thinning under Gaussian noise
to separate adaptive sampling from confidence interval construction,
achieving near-optimal cumulative regret through phased elimination.
RKHS regression can also serve as an estimation oracle within broader
sequential decision-making frameworks. Such oracle-based approaches
reduce contextual bandits to regression
\citep{foster2020beyond,simchi2020bypassing,hu2025eluder}
and reinforcement learning to regression
\citep{qian2024offline,hu2026model}.

\paragraph{The open problem and its partial answers.}
\citet{vakili2021open} posed the question of tight online confidence
intervals for RKHS elements. \citet{lattimore2023lower} resolved its
continuous-time version, with a lower bound for a direction that is not the
feature vector of any point of the domain and on a domain that grows with
\(n\), and left the discrete-time problem open. \citet{khamaru2021near}
carried out the discretization in finite dimension, where
\(\sqrt{\gamma_n}\) and the rate \(\sqrt{d\log n}\) coincide. On the
regret side, \citet{whitehouse2023sublinear}, who describe their result as a
partial resolution of the open problem, show that GP-UCB attains sublinear
regret for the Mat\'ern family when the regularization parameter grows with
the horizon. Our analysis establishes a discrete-time lower bound at a prespecified
point of a fixed compact domain for a kernel with polynomially growing
information gain. Together with the existing upper bound, this yields the
sharp minimax characterization in Corollary~\ref{cor:minimax}.

\section{Preliminaries}
\label{sec:setup}

In this section, we formalize the statistical model and the criterion used
throughout the paper. We first recall the RKHS model and the online
design protocol, following the notation of \citet{vakili2021open}. We then describe the surrogate GP model and the form of the confidence
intervals, state the existing upper bound, and finally define the minimax
confidence width.

\paragraph{Reproducing kernel Hilbert spaces.}
Let \(\mathcal X\subset\mathbb R^d\) be a compact set and let
\(k:\mathcal X\times\mathcal X\to\mathbb R\) be a continuous positive definite
kernel. Let \(\mathcal H_k\) denote the RKHS
corresponding to \(k\), defined as the Hilbert space of functions on
\(\mathcal X\) equipped with an inner product
\(\langle\cdot,\cdot\rangle_{\mathcal H_k}\) that satisfies
\[
k(\cdot,x)\in\mathcal H_k,
\ 
\langle f,k(\cdot,x)\rangle_{\mathcal H_k}=f(x),
\ 
\forall x\in\mathcal X,\ \forall f\in\mathcal H_k.
\]
The second identity is the reproducing property. In particular, we have
\(\|k(\cdot,x)\|_{\mathcal H_k}^2=k(x,x)\). Let
\(\{\lambda_m\}_{m=1}^\infty\) and \(\{\phi_m\}_{m=1}^\infty\) denote the
Mercer eigenvalues and eigenfeatures of \(k\) with respect to a finite Borel
measure whose support is \(\mathcal X\). By Mercer's representation theorem
\citep{kanagawa2018gaussian}, every
\(f\in\mathcal H_k\) can be written as
\begin{equation}
f(x)=\sum_{m=1}^\infty w_m\lambda_m^{1/2}\phi_m(x),
\ 
\|f\|_{\mathcal H_k}=\|w\|_{\ell_2},
\label{eq:mercer}
\end{equation}
so the decay of the eigenvalues \(\lambda_m\) measures the size of the
function class. We use two examples below. The linear kernel
\(k(x,x')=x^\top x'\) has at most \(d\) nonzero eigenvalues, and its RKHS
consists of linear functions. A kernel on the circle of the form
\(\sum_{l\ge1}\alpha_l\cos(l(b-b'))\), with \(\alpha_l\ge0\) and
\(\sum_{l\ge1}\alpha_l<\infty\), has the trigonometric functions as
eigenfeatures, and its eigenvalues are proportional to the weights
\(\alpha_l\), each with multiplicity two.

\paragraph{Observation model and online designs.}
A learner collects noisy observations \(\{(x_i,y_i)\}_{i=1}^n\) of an unknown
objective function \(f\in\mathcal H_k\), where
\[
y_i=f(x_i)+\epsilon_i,
\ i\in[n].
\]
This is the standard observation model for kernelized bandits
\citep{srinivas2010gaussian,chowdhury2017kernelized}.
Following \citet{vakili2021open}, we impose the following assumption.

\begin{assumption}
\label{ass:model}
The RKHS norm of \(f\) is bounded as \(\|f\|_{\mathcal H_k}\le B\), for some
\(B>0\). Moreover, the noise terms are i.i.d.\ sub-Gaussian random variables,
i.e., for some \(R>0\),
\(\mathbb E[\exp(\eta\epsilon_i)]\le\exp(\eta^2R^2/2)\) for all
\(\eta\in\mathbb R\) and all \(i\ge1\).
\end{assumption}

In online learning problems, as pointed out by \citet{vakili2021open}, the observation points are collected
sequentially: the point \(x_{i+1}\) is determined after all the values
\(\{(x_j,y_j)\}_{j=1}^i\) are revealed. This is in contrast to an offline
design, where the points are fixed in advance.

\begin{definition}
\label{def:online}
(i) \emph{Online setting.} Define the filtration
\[
\mathcal F_i
:= \sigma(x_1,\ldots,x_{i+1},\epsilon_1,\ldots,\epsilon_i),
\ i\ge0.
\]
For each \(i\ge1\), \(x_i\) is \(\mathcal F_{i-1}\)-measurable. The noise
\(\epsilon_i\) is \(\mathcal F_i\)-measurable and is independent of
\(\mathcal F_{i-1}\).

(ii) \emph{Offline setting.} The design sequence \((x_i)_{i\ge1}\) is
independent of the noise sequence \((\epsilon_i)_{i\ge1}\).
\end{definition}

We make two additions to part (i) explicit. First, we include the sigma
algebra \(\mathcal F_0=\sigma(x_1)\), which is needed for \(x_1\). Second, we
state the independence of \(\epsilon_i\) from \(\mathcal F_{i-1}\), which is
implicit in \citet{vakili2021open}: it holds whenever \(x_i\) is a measurable
function of the past observations \(\{(x_j,y_j)\}_{j<i}\) and of
randomization that is independent of the noise. When only \(n\) rounds are
considered, we impose these conditions for \(i\in[n]\). We refer to a
sequence \(\{x_i\}_{i=1}^n\) that satisfies part (i) as an
\emph{online design}.

\paragraph{Surrogate GP model.}
A standard approach in kernelized regression employs a zero-mean surrogate GP
model \(\hat f\) with covariance kernel \(k\), using its posterior mean to
estimate \(f\) and its posterior standard deviation to quantify uncertainty
\citep{srinivas2010gaussian,chowdhury2017kernelized,vakili2021optimal}.
GP model \(\hat f\) with kernel \(k\) provides a surrogate posterior
mean
\(\mu_n(x)=\mathbb E[\hat f(x)\mid\{(x_i,y_i)\}_{i=1}^n]\)
and a surrogate posterior variance
\(\sigma_n^2(x)=\mathbb E[(\hat f(x)-\mu_n(x))^2\mid\{(x_i,y_i)\}_{i=1}^n]\).
It is well known that
\begin{equation}
\mu_n(x)=z_n^\top(x)y_n,
\ 
\sigma_n^2(x)=k(x,x)-k_n^\top(x)(\lambda^2I_n+K_n)^{-1}k_n(x),
\label{eq:posterior}
\end{equation}
where \(k_n(x)=[k(x,x_1),k(x,x_2),\ldots,k(x,x_n)]^\top\), \(K_n\) is the
kernel matrix \([K_n]_{i,j}=k(x_i,x_j)\),
\(z_n(x)=(\lambda^2I_n+K_n)^{-1}k_n(x)\), and \(\lambda>0\) is a
regularization parameter.  \(y_n=[y_1,y_2,\ldots,y_n]^\top\) denotes the vector
of observations. The posterior mean \(\mu_n\) coincides with the kernel ridge
regression estimate with penalty \(\lambda^2\). The GP model is only a
surrogate: the objective function \(f\) is a fixed element of
\(\mathcal H_k\), and the noise need not be Gaussian.

\paragraph{Confidence intervals and the existing bound.}
A confidence interval for \(f\) takes the form
\begin{equation}
|f(x)-\mu_n(x)|\le\rho_n(\delta)\,\sigma_n(x)
\ \text{ for all }x\in\mathcal X,
\ \text{ with probability at least }1-\delta,
\label{eq:ci}
\end{equation}
and we refer to \(\rho_n(\delta)\) as its \emph{width}.

In the online setting, we use the maximal information gain
\begin{equation}
\gamma_n
:=
\sup_{(x_1,\ldots,x_n)\in\mathcal X^n}
\log\det\left(I_n+\lambda^{-2}K_n\right).
\label{eq:gamma}
\end{equation}
This quantity is closely related to the effective dimension of the kernel
\citep{srinivas2010gaussian,valko2013finite}. It is the normalized form of
the definition in the open problem \citep{vakili2021open}, since
\[
\log\det\left(I_n+\lambda^{-2}K_n\right)
=
\log\det\left(\lambda^2 I_n+K_n\right)-n\log(\lambda^2).
\]
The two definitions, therefore, have the same maximizing designs and agree
in value when \(\lambda=1\). For the linear kernel, we have
\(\gamma_n=O(d\log n)\),
whereas \(\gamma_n\) can grow polynomially in \(n\) for kernels whose
Mercer eigenvalues decay polynomially
\citep{vakili2021information,janz2026maximum}.

Several works study online confidence bounds for kernel regression
\citep{abbasi2013thesis,chowdhury2017kernelized,vakili2021open,whitehouse2023sublinear}.
We state the bound for any fixed \(\lambda>0\) in our notation.

\begin{theorem}[\citet{chowdhury2017kernelized,durand2018streaming}]
\label{thm:upper_bound}
Let \(k\) be a continuous positive definite kernel on a compact set
\(\mathcal X\subset\mathbb R^d\). Under Assumption~\ref{ass:model}, in the
online design setting of Definition~\ref{def:online}, for any \(\delta\in(0,1)\),
with probability at least \(1-\delta\), we have for all \(n\ge1\) and all
\(x\in\mathcal X\) that
\begin{equation}
|f(x)-\mu_n(x)|
\le
\left(
B+\frac{R}{\lambda}\sqrt{2\log\frac1\delta+\gamma_n}
\right)\sigma_n(x).
\label{eq:upper}
\end{equation}
\end{theorem}
Theorem~\ref{thm:upper_bound} holds for every kernel and every online design.

\paragraph{The minimax confidence width.}
To provide the minimax lower bound, we first define the notion of minimax confidence width.
We call a pair \((f,\mathbb P)\), consisting of an objective function \(f\)
and a joint law \(\mathbb P\) of the design and the noise
\(\{(x_i,\epsilon_i)\}_{i=1}^n\), an \emph{instance}, and we write
\(\mathcal P_n(B,R)\) for the collection of all instances that satisfy
Assumption~\ref{ass:model} and the online design condition in
Definition~\ref{def:online}. We define the minimax confidence width by
\begin{align}
\rho_n^*(\delta)
:=
\inf\Biggl\{\rho\ge0:
\sup_{(f,\mathbb P)\in\mathcal P_n(B,R)}
\mathbb P\Bigl(
\exists x\in\mathcal X:\,
|f(x)-\mu_n(x)|>\rho\,\sigma_n(x)
\Bigr)\le\delta
\Biggr\}.
\label{eq:minimax-width}
\end{align}
The quantity \(\rho_n^*(\delta)\) depends on the kernel, the domain, and
\((B,R,\lambda)\), which we suppress in the notation. The event in
\eqref{eq:minimax-width} is measurable, because \(f\), \(\mu_n\), and
\(\sigma_n\) are continuous in \(x\) and \(\mathcal X\) is separable. The
supremum takes the worst-case failure probability over the objective
function, the noise distribution, and the online design. The design may
depend on \(n\), \(B\), \(R\), and \(\lambda\). The outer infimum selects
the smallest width with worst-case failure probability at most \(\delta\).

Theorem~\ref{thm:upper_bound} gives
\[
\rho_n^*(\delta)
\le B+\frac{R}{\lambda}\sqrt{2\log(1/\delta)+\gamma_n}
\]
for every kernel and every \(n\). The open problem \citep{vakili2021open} asks how
\(\rho_n^*(\delta)\) grows with \(n\), and in particular whether the
\(\sqrt{\gamma_n}\) dependence can be improved.

\section{The Sharp Minimax Confidence Width}
\label{sec:thms}

In this section, we provide our theoretical guarantees on the sharp minimax
confidence width. We first specify the kernel and the online design of the
hard instance and establish a lower bound for online confidence intervals. Our additive Fourier kernel and adaptive sampling rule are partially motivated by
the construction of \citet{lattimore2023lower}.
We then combine this result with the existing upper bound of
Theorem~\ref{thm:upper_bound} to obtain a sharp characterization of the
minimax confidence width, and we record the resulting answer to the open
problem.

\paragraph{The hard kernel.}
Fix a smoothness parameter \(s>1/2\). Define the \(2\pi\)-periodic function
\[
\chi(\vartheta):=\sum_{l=1}^\infty\alpha_l\cos(l\vartheta).
\]
Let \(\zeta(u):=\sum_{m=1}^\infty m^{-u}\) for \(u>1\) denote the Riemann
zeta function. The weights are \(\alpha_l:=l^{-2s}/\zeta(2s)\) for
\(l\ge1\), so \(\sum_{l\ge1}\alpha_l=1\).

Using the Fourier representation of periodic RKHS kernels in
\citet{krieg2023periodic}, we define the normalized periodic Sobolev
kernel of order \(s\) on \([0,2\pi]\) by
\begin{equation}
k_s^\circ(b,b'):=\chi(b-b').
\label{eq:sobolev-kernel}
\end{equation}
The normalization gives \(k_s^\circ(b,b)=\chi(0)=1\). Under the uniform
distribution on \([0,2\pi]\), the eigenfunctions are
\(\sqrt{2}\cos(lb)\) and \(\sqrt{2}\sin(lb)\) for \(l\ge1\), each with
eigenvalue \(\alpha_l/2\). The associated RKHS is the mean-zero periodic
Sobolev space of order \(s\) \citep{wahba1990spline,krieg2023periodic}.

Our second kernel is the linear kernel on \([-1,1]\), given by
\[
k_2(a,a'):=aa',\  a,a'\in[-1,1].
\]

For the hard instance, we take the domain
\(\mathcal X:=[-1,1]\times[0,2\pi]\subset\mathbb R^2\) and combine the
two kernels to obtain
\begin{equation}
k\bigl((a,b),(a',b')\bigr):=k_2(a,a')+k_s^\circ(b,b').
\label{eq:lifted-kernel}
\end{equation}
The constructed kernel satisfies the following property.

\begin{proposition}
\label{prop:hard-kernel}
The kernel \(k\) in \eqref{eq:lifted-kernel} is continuous and positive
definite, with \(k((a,b),(a,b))=1+a^2\le2\). Under the uniform probability
measure on \(\mathcal X\), its nonzero Mercer eigenvalues consist of one
copy of \(1/3\) and two copies of \(\alpha_l/2\) for each \(l\ge1\).
When listed in nonincreasing order with multiplicity, these eigenvalues
satisfy \(\lambda_m\asymp m^{-2s}\).
\end{proposition}

The proof is given in Appendix~\ref{app:hard-kernel}.

\paragraph{Data generation.}
Fix \(n\ge40\), set \(f_B(a,b):=-Ba\), and let \(q:=\lfloor n/20\rfloor\).
Before sampling, we fix the second coordinates by visiting \(q\) equally
spaced points in the same order for \(20\) passes, then using \(0\) for
the remaining rounds.
\begin{equation}
b_{(r-1)q+j}:=\frac{2\pi(j-1)}{q}
\ \text{for } r\in[20],\ j\in[q],
\ 
b_i:=0\ \text{for }20q<i\le n.
\label{eq:cyclic-design}
\end{equation}
From these coordinates, we form \(K_n^\circ:=[k_s^\circ(b_i,b_j)]_{i,j\in[n]}\)
and compute the inverse and the \(LDL^\top\) factorization of the
regularized kernel matrix.
\begin{equation}
G:=(\lambda^2I_n+K_n^\circ)^{-1},
\ 
\lambda^2I_n+K_n^\circ=LDL^\top,
\label{eq:GLD}
\end{equation}
where \(L\) is unit lower triangular and \(D\) is diagonal with positive
entries. Set
\begin{equation}
E_n:=\tr(G)-\sum_{i=1}^n\frac{1}{D_{ii}},
\ 
\kappa:=E_n^{-1/2}.
\label{eq:En}
\end{equation}
The repeated visit to \(b_1\) ensures \(E_n>0\), as shown in the proof of
Theorem~\ref{thm:lower_bound}.

In round \(i\in[n]\), take \(b_i\) from \eqref{eq:cyclic-design} and choose
\(x_i\) using the past residuals \(y_j+B\bar a_j\), \(j<i\).
\begin{equation}
a_i:=-\frac{\kappa}{R}\sum_{j<i}L_{ij}(y_j+B\bar a_j),
\ 
\bar a_i:=\max\{-1,\min\{1,a_i\}\},
\ 
x_i:=(\bar a_i,b_i).
\label{eq:online-design}
\end{equation}
We then observe \(y_i=f_B(x_i)+\epsilon_i=-B\bar a_i+\epsilon_i\), where
\(\epsilon_i\sim\mathcal N(0,R^2)\) is independent of the past.
We evaluate the confidence interval at the fixed point \(x_0:=(1,0)\).

\begin{theorem}[Lower bound for online confidence intervals]
\label{thm:lower_bound}
Fix \(s>1/2\) and \(B,R,\lambda>0\). Consider the kernel
\eqref{eq:lifted-kernel}, the objective function \(f_B(a,b)=-Ba\), i.i.d.\ noise
\(\epsilon_i\sim\mathcal N(0,R^2)\), and the online design
\eqref{eq:cyclic-design}--\eqref{eq:online-design}. There exist constants
\(c_s,c_s'>0\) depending only on \(s\) and an integer
\(n_0=n_0(s,\lambda)\ge\lambda^2\) such that, for every \(n\ge n_0\), with
probability at least \(6/11\),
\[
\frac{|f_B(x_0)-\mu_n(x_0)|}{\sigma_n(x_0)}
\ge
c_s\left[B+\frac{R}{\lambda}\Big(\frac{n}{\lambda^2}\Big)^{\frac{1}{4s}}\right]
\ge
c_s'\left(B+\frac{R}{\lambda}\sqrt{\gamma_n}\right).
\]
Moreover, for every \(n\ge n_0\), the maximal information gain of the kernel
\eqref{eq:lifted-kernel} satisfies
\(\gamma_n\asymp(n/\lambda^2)^{1/(2s)}\), with comparison constants depending
only on \(s\).

Consequently, for every \(\delta\in(0,1/2]\) and \(n\ge n_0\), every
deterministic width \(\rho_n(\delta)\) for which \eqref{eq:ci} holds for
this instance satisfies
\[
\rho_n(\delta)
\ge
c_s'\left(B+\frac{R}{\lambda}\sqrt{\gamma_n}\right).
\]
\end{theorem}

Combining the upper bound in Theorem~\ref{thm:upper_bound} and the lower
bound in Theorem~\ref{thm:lower_bound} with a Gaussian tail lower bound for
a deterministic design gives the sharp minimax confidence width.

\begin{corollary}[Sharp minimax confidence width]
\label{cor:minimax}
Fix \(s>1/2\) and \(\lambda>0\), and consider the kernel
\eqref{eq:lifted-kernel}. Let \(n_0=n_0(s,\lambda)\) be as in
Theorem~\ref{thm:lower_bound}. For all \(n\ge n_0\), \(B,R>0\), and
\(\delta\in(0,1/4]\), we have
\[
\rho_n^*(\delta)
\asymp
B+\frac{R}{\lambda}\sqrt{\gamma_n+\log\frac1\delta},
\]
with comparison constants depending only on \(s\).
\end{corollary}
For fixed \(s,B,R,\lambda\) and \(\delta\in(0,1/4]\), the minimax width
for our \(d=2\) kernel grows as \(n^{1/(4s)}\), with the lower bound
holding even at the prespecified point \(x_0\).
Therefore, we resolve the open problem posed by \citet{vakili2021open}
in the worst case over kernels in discrete time.

\section{Discussion}
\label{sec:discussion}

This paper resolves, in the worst case over kernels, the open problem on
tight online confidence intervals for RKHS elements posed by
\citet{vakili2021open}, in the discrete-time model in which it was stated. We use the existing upper bound and establish a matching
lower bound. In particular, for the additive kernel
\eqref{eq:lifted-kernel}, combining our lower bound with the existing
self-normalized upper bound
\citep{abbasi2013thesis,chowdhury2017kernelized,durand2018streaming} yields
\[
\rho_n^*(\delta)
\asymp
B+\frac{R}{\lambda}\sqrt{\gamma_n+\log\frac1\delta}
\]
for all sufficiently large \(n\) and all \(\delta\in(0,1/4]\). Our  discrete-time analysis shows that, for intervals of the
form \eqref{eq:ci} and in the worst case over kernels and online designs,
this inflation is an unavoidable statistical obstruction rather than an
artifact of a particular concentration inequality or proof technique. Thus, the \(\sqrt{\gamma_n}\) dependence of the existing confidence width
is unavoidable in general, ruling out the conjectured polylogarithmic
width. It is also consistent with the fact that the known
order-optimal algorithms for kernelized bandits rely on batching or
elimination, which control the dependence between the design and the noise,
rather than on a sharper online interval.

Several directions follow from this characterization. The most immediate task
is to determine the minimax confidence width for specific kernel families.
Our lower bound uses an additive linear component, and the online design
steers the estimate of its coefficient. It remains open whether a Mat\'ern or
squared exponential kernel alone exhibits the same inflation at a point \citep{janz2026maximum}. A second direction is the regret of GP-UCB with a
fixed regularization parameter, for which our result does not exclude a
design-specific analysis. A third direction is a
lower bound over all estimators, as in the Bayes-risk bound of
\citet{lattimore2023lower} and the minimax bound of
\citet{khamaru2021near}, and over all interval constructions, for a fixed
kernel with polynomially growing information gain. Finally, a natural direction is to construct data-driven widths that adapt
to the realized design while maintaining coverage. On the hard instance of Theorem~\ref{thm:lower_bound}, any data-dependent
interval
\(\mu_n(x_0)\pm\widehat\rho_n(\delta)\sigma_n(x_0)\)
with coverage at least \(1-\delta\) must satisfy
\(\widehat\rho_n(\delta)\ge c_s'[B+(R/\lambda)\sqrt{\gamma_n}]\)
with probability at least \(6/11-\delta\), for every \(n\ge n_0\).
Wild refitting and related resampling methods provide tools for
evaluating excess risk and prediction error
\citep{wainwright2025wild,hu2025perturbing,hu2026interleaved,ni2026upper}.
An open question is whether these ideas can be adapted to pointwise
confidence intervals under adaptive sampling. We leave the investigation of these directions to future work.

\bibliographystyle{plainnat}
\bibliography{refs}
\appendix
\clearpage
\section{Useful Mathematical Tools}
\label{app:tools}

We collect the concentration inequality and matrix formulas used below.

\begin{lemma}[Laurent--Massart inequality, \citet{laurent2000adaptive}]
\label{lem:LM}
Let $\xi\sim\mathcal N(0,I_n)$ and let $A\in\mathbb R^{n\times n}$ be
symmetric and positive semidefinite. For every $t>0$,
\begin{align*}
\mathbb P\!\left(\xi^\top A\xi>
\operatorname{tr}(A)+2\|A\|_F\sqrt t+2\|A\|_{\mathrm{op}}t\right)
&\le e^{-t},\\
\mathbb P\!\left(\xi^\top A\xi<
\operatorname{tr}(A)-2\|A\|_F\sqrt t\right)
&\le e^{-t}.
\end{align*}
\end{lemma}

\begin{lemma}[Rank-one updates, \citet{horn2012matrix}]
\label{lem:rank-one-update}
Let $A\in\mathbb R^{n\times n}$ be symmetric positive definite and
$u\in\mathbb R^n$. Then
\begin{align*}
(A+uu^\top)^{-1}
&=A^{-1}-\frac{A^{-1}uu^\top A^{-1}}{1+u^\top A^{-1}u},\\
\det(A+uu^\top)&=\det(A)(1+u^\top A^{-1}u).
\end{align*}
\end{lemma}

\begin{lemma}[Schur complements, \citet{horn2012matrix}]
\label{lem:schur-complement}
Let
$A=\begin{pmatrix}P&q\\q^\top&r\end{pmatrix}\in\mathbb R^{n\times n}$
be symmetric positive definite. Its Schur complement
$r-q^\top P^{-1}q$ is positive, and
\[
\det A=\det P\,(r-q^\top P^{-1}q),
\ (A^{-1})_{nn}=(r-q^\top P^{-1}q)^{-1}.
\]
For $n=1$, the empty determinant and quadratic form are interpreted as
one and zero, respectively.
\end{lemma}

\begin{lemma}[Determinant identities and bounds, \citet{horn2012matrix}]
\label{lem:determinant-tools}
For $\Psi\in\mathbb R^{n\times m}$, Sylvester's determinant identity gives
\[
\det(I_n+\Psi\Psi^\top)=\det(I_m+\Psi^\top\Psi).
\]
For every symmetric positive definite $A\in\mathbb R^{m\times m}$,
Hadamard's inequality gives
\[
\det A\le\prod_{j=1}^m A_{jj}.
\]
\end{lemma}

\section{Properties of the Hard Kernel}
\label{app:hard-kernel}

\begin{proof}[Proof of Proposition~\ref{prop:hard-kernel}]
Since $s>1/2$, the weights satisfy $\sum_{l\ge1}\alpha_l=1$.
The Fourier series defining $\chi$ therefore converges uniformly, so
$k_s^\circ$ and the additive kernel $k$ are continuous.
Define
\[
e_0(a,b):=\sqrt3\,a,
\ e_{l,c}(a,b):=\sqrt2\cos(lb),
\ e_{l,s}(a,b):=\sqrt2\sin(lb),\ l\ge1.
\]
The cosine difference identity gives the uniformly convergent expansion
\[
k(x,x')
=\frac13e_0(x)e_0(x')
+\sum_{l=1}^\infty\frac{\alpha_l}{2}
\{e_{l,c}(x)e_{l,c}(x')+e_{l,s}(x)e_{l,s}(x')\}.
\]
Each feature product is a positive semidefinite rank-one kernel, and
all its coefficients are positive. We conclude that $k$ is positive definite. Setting $x=x'=(a,b)$
and using $\cos^2(lb)+\sin^2(lb)=1$ yields
$k(x,x)=a^2+\sum_{l\ge1}\alpha_l=1+a^2\le2$.

Under the uniform probability measure on $\mathcal X$, the functions
$e_0,e_{l,c},e_{l,s}$ form an orthonormal family. The trigonometric
functions are orthonormal under the uniform measure on $[0,2\pi]$.
The first coordinate is independent of the second, with mean zero and
second moment $1/3$. Integrating the kernel expansion against each of
these functions gives eigenvalue $1/3$ for $e_0$ and $\alpha_l/2$
for both $e_{l,c}$ and $e_{l,s}$. The integral operator vanishes on
the orthogonal complement of their closed span, so these are all its
nonzero eigenvalues.

The periodic eigenvalues in nonincreasing order are
$\tilde\lambda_m=(2\zeta(2s))^{-1}\lceil m/2\rceil^{-2s}$.
After inserting the single eigenvalue $1/3$, the ordered sequence satisfies
$\tilde\lambda_m\le\lambda_m\le\tilde\lambda_{m-1}$ for $m\ge2$.
Since $\tilde\lambda_m\asymp m^{-2s}$, we obtain the claimed decay.
\end{proof}

\section{Proof of the Lower Bound}
\label{app:lower-bound}

Fix a positive semidefinite kernel $k^\circ$ on $[0,2\pi]$ with
$k^\circ(b,b)\le1$, a deterministic sequence $b_1,\ldots,b_n$, and a test
point $x_0=(1,b_0)$ with $b_0\in[0,2\pi]$. We use the additive kernel
$k((a,b),(a',b'))=aa'+k^\circ(b,b')$ and the response model
$y_i=f_B(x_i)+\epsilon_i$ with $f_B(a,b)=-Ba$ and i.i.d.\
$\epsilon_i\sim\mathcal N(0,R^2)$. Write $K_n^\circ=[k^\circ(b_i,b_j)]_{i,j\in[n]}$
and define $G,L,D,E_n$ as in \eqref{eq:GLD} and \eqref{eq:En}. For
$J\subseteq[n]$, let $k_J^\circ(b)=(k^\circ(b,b_j))_{j\in J}$ and let
$K_J^\circ$ be the corresponding principal submatrix of $K_n^\circ$. The
posterior variance for the $k^\circ$ component based on $J$ is
\[
v_J(b):=k^\circ(b,b)-k_J^\circ(b)^\top(\lambda^2I_{|J|}+K_J^\circ)^{-1}k_J^\circ(b),
\ v_\emptyset(b):=k^\circ(b,b).
\]
Write $v_{<i}=v_{[i-1]}$ and $v_{-i}=v_{[n]\setminus\{i\}}$.

Whenever $E_n>0$, set $\kappa=E_n^{-1/2}$ and use the design
\eqref{eq:online-design}. The residual $y_i+B\bar a_i$ equals $\epsilon_i$
for every realization, including when $a_i$ is clipped. Thus the vector
$\xi$ with coordinates $\xi_i=(y_i+B\bar a_i)/R=\epsilon_i/R$ has law
$\mathcal N(0,I_n)$. The offset vector before clipping is
\[
\tilde a:=(I_n-L)\xi,
\ a=\kappa\tilde a.
\]
Indeed, $I_n-L$ is strictly lower triangular, so
$a_i=-\kappa\sum_{j<i}L_{ij}\xi_j$ depends only on past residuals. The actual
point is $x_i=(\bar a_i,b_i)$.

\subsection{The posterior error at the test point}

For vectors $a,\xi\in\RR^n$, define
\[
\Gamma:=a^\top G\xi,
\ Q:=a^\top Ga,
\ \theta:=k_{[n]}^\circ(b_0)^\top Ga,
\ \psi:=k_{[n]}^\circ(b_0)^\top G\xi.
\]

\begin{lemma}\label{lem:dual}
Suppose that $a\in[-1,1]^n$, $x_i=(a_i,b_i)$, and $y_n=R\xi-Ba$.
Then
\begin{align*}
\mu_n(x_0)-f_B(x_0)&=\frac{(1-\theta)(B+R\Gamma)}{1+Q}+R\psi,\\
\sigma_n^2(x_0)&=\frac{(1-\theta)^2}{1+Q}+v_{[n]}(b_0).
\end{align*}
\end{lemma}
\begin{proof}[Proof of Lemma~\ref{lem:dual}]
Let $h=k_{[n]}^\circ(b_0)$. The additive kernel gives
$K_n=K_n^\circ+aa^\top$, $k_n(x_0)=a+h$, and $k(x_0,x_0)=1+k^\circ(b_0,b_0)$.
Since $G\succ0$, we have $1+Q\ge1$. Applying Lemma~\ref{lem:rank-one-update}, we obtain
\[
A:=(\lambda^2I_n+K_n)^{-1}=G-\frac{Gaa^\top G}{1+Q}.
\]
Using this inverse, we obtain $(a+h)^\top Aa=(Q+\theta)/(1+Q)$ and
\[
(a+h)^\top A\xi=\Gamma+\psi-\frac{(Q+\theta)\Gamma}{1+Q}
=\frac{(1-\theta)\Gamma}{1+Q}+\psi.
\]
Substituting $y_n=R\xi-Ba$ into \eqref{eq:posterior} and subtracting
$f_B(x_0)=-B$ yields the claimed error. For the variance, we obtain
\[
\sigma_n^2(x_0)
=v_{[n]}(b_0)+1-Q-2\theta+\frac{(Q+\theta)^2}{1+Q}
=v_{[n]}(b_0)+\frac{(1-\theta)^2}{1+Q}.
\]
\end{proof}

\subsection{Cholesky identities and Gaussian bounds}

\begin{lemma}\label{lem:cholesky}
For $J\subseteq J'\subseteq[n]$ and $b\in[0,2\pi]$, we have
$0\le v_{J'}(b)\le v_J(b)\le k^\circ(b,b)$. For each $i\in[n]$,
\begin{equation}
D_{ii}=\lambda^2+v_{<i}(b_i)\ge\lambda^2,
\ G_{ii}=\frac{1}{\lambda^2+v_{-i}(b_i)}.
\label{eq:chol-diag}
\end{equation}
Consequently,
\begin{equation}
\begin{aligned}
E_n
&=\sum_{i=1}^n
\left[\frac{1}{\lambda^2+v_{-i}(b_i)}-\frac{1}{\lambda^2+v_{<i}(b_i)}\right]=\frac1{\lambda^2}\sum_{i=1}^n
\left[\frac{v_{<i}(b_i)}{\lambda^2+v_{<i}(b_i)}
-\frac{v_{-i}(b_i)}{\lambda^2+v_{-i}(b_i)}\right],
\end{aligned}
\label{eq:En-variance-form}
\end{equation}
with every summand nonnegative.
\end{lemma}
\begin{proof}[Proof of Lemma~\ref{lem:cholesky}]
Completing the square, we obtain
\[
v_J(b)=\min_{z\in\RR^{|J|}}
\left[k^\circ(b,b)-2z^\top k_J^\circ(b)
+z^\top(\lambda^2I_{|J|}+K_J^\circ)z\right].
\]
The expression is nonnegative by positive semidefiniteness of the Gram matrix
on the points in $J$ and $b$. Extending $z$ by zero on $J'\setminus J$ shows
$v_{J'}(b)\le v_J(b)$. The choice $z=0$ gives $v_J(b)\le k^\circ(b,b)$.

The leading $i$ by $i$ principal submatrix of $LDL^\top$ has determinant
$\prod_{j\le i}D_{jj}$ because $L$ is unit lower triangular. Taking the ratio
of the leading determinants of orders $i$ and $i-1$ gives $D_{ii}$, with
the empty determinant interpreted as one. Applying
Lemma~\ref{lem:schur-complement} to that leading submatrix, we evaluate
this ratio as $\lambda^2+v_{<i}(b_i)$. Applying the same lemma to the full
matrix after moving index $i$ to the last position gives
$G_{ii}=(\lambda^2+v_{-i}(b_i))^{-1}$.
Substituting both identities into \eqref{eq:En} gives the first sum in
\eqref{eq:En-variance-form}, whose summands are nonnegative since
$v_{-i}(b_i)\le v_{<i}(b_i)$. Rewriting
$\lambda^2/(\lambda^2+v)=1-v/(\lambda^2+v)$ gives the second sum.
\end{proof}

Define $H:=D^{-1/2}L^{-1}$ and $M:=H-D^{-1/2}$. For $\xi\sim\mathcal N(0,I_n)$,
write
\[
u:=M\xi,
\ W:=\|u\|_2^2,
\ Z:=\sum_{i=1}^n u_iD_{ii}^{-1/2}\xi_i.
\]

\begin{lemma}\label{lem:structure}
The matrix $M$ is strictly lower triangular, so $u_i$ depends only on
$\xi_1,\ldots,\xi_{i-1}$. For the offset vector $a=\kappa(I_n-L)\xi$,
\[
\Gamma=\kappa(W+Z),
\ Q=\kappa^2W,
\ \EE[W]=\|M\|_F^2=E_n.
\]
\end{lemma}
\begin{proof}[Proof of Lemma~\ref{lem:structure}]
Since $L^{-1}$ is unit lower triangular, $M=D^{-1/2}(L^{-1}-I_n)$ is
strictly lower triangular. We have $G=H^\top H$ and
\[
Ha=\kappa u,
\ H\xi=u+D^{-1/2}\xi.
\]
We pair $Ha$ with $H\xi$ and with itself to obtain the identities for
$\Gamma$ and $Q$. Since $\EE[\xi\xi^\top]=I_n$, we have $\EE[W]=\|M\|_F^2$.
The strictly lower triangular part $M$ and the diagonal part $D^{-1/2}$
of $H$ are orthogonal under the Frobenius inner product. Therefore,
\[
\|M\|_F^2=\|H\|_F^2-\sum_{i=1}^nD_{ii}^{-1}
=\tr(G)-\sum_{i=1}^nD_{ii}^{-1}=E_n.
\]
\end{proof}

\begin{lemma}\label{lem:variance}
For the offset vector before clipping, each $\tilde a_i$, $\psi$, and
$\theta$ is centered Gaussian, with
\begin{equation}
\Var(\tilde a_i)\le\frac{1}{\lambda^2},
\ \Var(\psi)\le\frac{1}{\lambda^2},
\ \Var(\theta)\le\frac{4\kappa^2}{\lambda^2}.
\label{eq:var-bounds}
\end{equation}
Moreover,
\begin{equation}
\|M^\top M\|_{\mathrm{op}}\le\frac4{\lambda^2},
\ \|M^\top M\|_F^2\le\frac{4E_n}{\lambda^2},
\ \sum_{i=1}^n\frac{u_i^2}{D_{ii}}\le\frac{W}{\lambda^2}.
\label{eq:norm-bounds}
\end{equation}
\end{lemma}
\begin{proof}[Proof of Lemma~\ref{lem:variance}]
The three scalars in \eqref{eq:var-bounds} are deterministic linear
functions of $\xi$. To bound the variance of $\tilde a_i$, use
\eqref{eq:chol-diag} and the diagonal of $LDL^\top$ to obtain
\[
\lambda^2\Var(\tilde a_i)=\lambda^2\sum_{j<i}L_{ij}^2
\le\sum_{j<i}D_{jj}L_{ij}^2=k^\circ(b_i,b_i)-v_{<i}(b_i)\le1.
\]
Since $K_n^\circ\succeq0$, we have
$\|H\|_{\mathrm{op}}^2=\|G\|_{\mathrm{op}}\le\lambda^{-2}$.
Combining this estimate with $D_{ii}\ge\lambda^2$ gives
$\|M\|_{\mathrm{op}}\le\|H\|_{\mathrm{op}}+\|D^{-1/2}\|_{\mathrm{op}}
\le2/\lambda$.

Let $h=k_{[n]}^\circ(b_0)$. Then
$\|Hh\|_2^2=h^\top Gh=k^\circ(b_0,b_0)-v_{[n]}(b_0)\le1$.
Using $G=H^\top H$ and $Ha=\kappa M\xi$, we obtain
\[
\Var(\psi)=\|H^\top Hh\|_2^2\le\lambda^{-2},
\ \Var(\theta)=\kappa^2\|M^\top Hh\|_2^2\le4\kappa^2\lambda^{-2}.
\]
Finally, since $M^\top M$ is positive semidefinite,
\[
\|M^\top M\|_F^2
\le\|M^\top M\|_{\mathrm{op}}\tr(M^\top M)
\le\frac{4E_n}{\lambda^2},
\]
where the last inequality uses Lemma~\ref{lem:structure} and the operator
bound for $M$. The last bound in \eqref{eq:norm-bounds} follows from
$D_{ii}\ge\lambda^2$ and $W=\sum_i u_i^2$.
\end{proof}

\subsection{A probability bound for the normalized error}

\begin{lemma}\label{lem:engine}
Fix $\delta_0\in(0,1/5)$ and suppose that
\begin{equation}
E_n\ge\underline E(\lambda,\delta_0,n)
:=\max\left\{
\frac{6272}{\lambda^2}\log\frac{2}{\delta_0},
\frac{2}{\lambda^2}\log\frac{2n}{\delta_0}
\right\}.
\label{eq:threshold}
\end{equation}
Under the design \eqref{eq:online-design}, with probability at least
$1-5\delta_0$,
\begin{equation}
\frac{|f_B(x_0)-\mu_n(x_0)|}{\sigma_n(x_0)}
\ge\frac{4B}{21\sqrt3}+R\sqrt{\frac{E_n}{9408}},
\ \sigma_n^2(x_0)\ge\frac{2}{21}.
\label{eq:engine}
\end{equation}
\end{lemma}
\begin{proof}[Proof of Lemma~\ref{lem:engine}]
Set $t=\log(2/\delta_0)$. The quadratic form $W=\xi^\top M^\top M\xi$ has
mean $E_n$. Applying Lemma~\ref{lem:LM} to $M^\top M$, we obtain,
with probability at least
$1-2e^{-t}=1-\delta_0$,
\[
|W-E_n|
\le 2\|M^\top M\|_F\sqrt t+2\|M^\top M\|_{\mathrm{op}}t
\le \frac4\lambda\sqrt{E_nt}+\frac{8t}{\lambda^2}
\le\frac{5E_n}{8}.
\]
Here the norm bounds are from Lemma~\ref{lem:variance}, and
$E_n\ge64t/\lambda^2$ makes the last two terms at most $E_n/2$ and $E_n/8$.
Denote the resulting event $3E_n/8\le W\le13E_n/8$ by $\mathcal A_W$.

For the martingale term, let $\mathcal G_i=\sigma(\xi_1,\ldots,\xi_i)$ and
$\iota_i=u_iD_{ii}^{-1/2}$. The coefficient $\iota_i$ is
$\mathcal G_{i-1}$-measurable by Lemma~\ref{lem:structure}, and
$\xi_i$ is standard Gaussian independent of $\mathcal G_{i-1}$.
For each sign $r\in\{-1,1\}$, let
\[
X_i^{(r)}
:=\exp\left(
\frac{r\lambda^2}{2}\sum_{j\le i}\iota_j\xi_j
-\frac{\lambda^4}{8}\sum_{j\le i}\iota_j^2
\right),
\ X_0^{(r)}=1.
\]
Conditional on $\mathcal G_{i-1}$, the Gaussian moment generating function
gives $\EE[\exp(r\lambda^2\iota_i\xi_i/2-\lambda^4\iota_i^2/8)
\mid\mathcal G_{i-1}]=1$. By nonnegativity and induction, each $X_i^{(r)}$
is integrable with mean one and
$\EE[X_i^{(r)}\mid\mathcal G_{i-1}]=X_{i-1}^{(r)}$.
On $\mathcal A_W\cap\{|Z|>W/2\}$, the sign $r$ of $Z$ satisfies
\[
\log X_n^{(r)}
\ge\frac{\lambda^2W}{4}-\frac{\lambda^4}{8}\frac{W}{\lambda^2}
=\frac{\lambda^2 W}{8}
\ge\frac{3\lambda^2E_n}{64},
\]
where we used $\sum_i\iota_i^2\le W/\lambda^2$ from
\eqref{eq:norm-bounds}. Applying Markov's inequality for both signs gives
\[
\PP(\mathcal A_W\cap\{|Z|>W/2\})
\le 2e^{-3\lambda^2E_n/64}\le\delta_0
\]
by \eqref{eq:threshold}.

Using the Gaussian variance bounds in \eqref{eq:var-bounds} and
$\kappa^2=E_n^{-1}$, we obtain
\begin{align*}
\PP(|\theta|>1/2)&\le 2e^{-\lambda^2E_n/32}\le\delta_0,\\
\PP(|\psi|>\lambda^{-1}\sqrt{2t})&\le 2e^{-t}=\delta_0,\\
\PP\left(\max_{i\in[n]}|a_i|>1\right)
&\le 2n e^{-\lambda^2E_n/2}\le\delta_0.
\end{align*}
The last line uses $\Var(a_i)\le1/(\lambda^2E_n)$ and a union bound.
The first and last lines use the two terms of \eqref{eq:threshold},
respectively.

Together these bounds give an event of probability at least $1-5\delta_0$
on which $\mathcal A_W$ holds, $|Z|\le W/2$, $|\theta|\le1/2$,
$|\psi|\le\lambda^{-1}\sqrt{2t}$, and no offset is clipped.
On this event the raw response vector is $y_n=R\xi-Ba$, so
Lemma~\ref{lem:dual} applies to the realized design. By
Lemma~\ref{lem:structure}, we have
\[
1+Q=1+W/E_n\in[11/8,21/8],
\ \Gamma\ge\kappa W/2\ge 3\sqrt{E_n}/16.
\]
The threshold gives $|\psi|\le\lambda^{-1}\sqrt{2t}\le\sqrt{E_n}/56$.
Since $1-\theta\in[1/2,3/2]$, the signed posterior error satisfies
\[
\mu_n(x_0)-f_B(x_0)
\ge\frac4{21}\left(B+\frac{3R\sqrt{E_n}}{16}\right)-R|\psi|
\ge\frac{4B}{21}+\frac{R\sqrt{E_n}}{56}.
\]
Using $0\le v_{[n]}(b_0)\le1$ in the variance identity, we obtain
\[
\frac2{21}
\le\sigma_n^2(x_0)
\le\frac{(3/2)^2}{11/8}+1=\frac{29}{11}<3.
\]
Dividing the error bound by $\sigma_n(x_0)$ and using $56^2\cdot3=9408$
proves \eqref{eq:engine}.
\end{proof}

\subsection{Spectral estimates for the cyclic design}
\label{app:spectral}

\begin{lemma}\label{lem:cyclic}
Let \(k^\circ=k_s^\circ\) with \(s>1/2\), and let \(q\ge2\) and \(p\ge3\)
be integers with \(n\ge pq\). Suppose that
\[
b_{(r-1)q+j}=\frac{2\pi(j-1)}{q},
\ r\in[p],\ j\in[q],
\]
with arbitrary \(b_i\in[0,2\pi]\) for \(pq<i\le n\). Let
\(\omega_0,\ldots,\omega_{q-1}\) be the eigenvalues of
\(K_q^\circ:=[k_s^\circ(b_j,b_{j'})]_{j,j'\in[q]}\), and let
\(H_{p-1}:=\sum_{r=1}^{p-1}1/r\). Then
\[
\lambda^2E_n
\ge\frac{H_{p-1}-1}{16}\,\abr{\cbr{l:\omega_l\ge\lambda^2}},
\ 
\abr{\cbr{l:\omega_l\ge\lambda^2}}
\ge\min\cbr{q-1,\ \floor{\rbr{\frac{q}{2\zeta(2s)\lambda^2}}^{1/(2s)}}}.
\]
\end{lemma}

\begin{proof}[Proof of Lemma~\ref{lem:cyclic}]
Since \([K_q^\circ]_{j,j'}=\chi(2\pi(j-j')/q)\), the matrix
\(K_q^\circ\) is circulant. Let \(J\) consist of \(r\) complete passes.
Ordering the indices of \(J\) by pass gives
\(K_J^\circ=\one_r\one_r^\top\otimes K_q^\circ\) and
\(k_J^\circ(b_j)=\one_r\otimes K_q^\circ e_j\), where \(e_j\) is
the \(j\)-th standard basis vector. Since
\((\one_r\one_r^\top)\one_r=r\one_r\), we have
\[
(\lambda^2I_{rq}+K_J^\circ)^{-1}k_J^\circ(b_j)
=\one_r\otimes\rbr{(\lambda^2I_q+rK_q^\circ)^{-1}K_q^\circ e_j}.
\]
Using this identity in the posterior variance and simplifying gives
\begin{equation}
v_J(b_j)
=\sbr{\lambda^2K_q^\circ(\lambda^2I_q+rK_q^\circ)^{-1}}_{jj}
=\frac{\lambda^2}{q}\sum_{l=0}^{q-1}\frac{\omega_l}{\lambda^2+r\omega_l}
=:\lambda^2\beta_r.
\label{eq:beta-r}
\end{equation}
The second equality uses the constant diagonal of the circulant matrix in
brackets. Since each \(\omega_l\ge0\), \(\beta_r\) is nonincreasing in
\(r\), with \(0\le\beta_r\le\beta_1\le1\).

For \(i=(r-1)q+j\) with \(r\le p-1\), the observations before \(i\) are
contained in the first \(r\) passes, and those with indices other than
\(i\) include all \(p-1\) passes other than pass \(r\). Monotonicity from
Lemma~\ref{lem:cholesky} and \eqref{eq:beta-r} therefore give
\(v_{<i}(b_i)\ge\lambda^2\beta_r\) and
\(v_{-i}(b_i)\le\lambda^2\beta_{p-1}\).
The \(i\)-th summand of \(\lambda^2E_n\) in \eqref{eq:En-variance-form}
is thus at least
\[
\frac{\beta_r}{1+\beta_r}-\frac{\beta_{p-1}}{1+\beta_{p-1}}
=\frac{\beta_r-\beta_{p-1}}{(1+\beta_r)(1+\beta_{p-1})}
\ge\frac{\beta_r-\beta_{p-1}}{4}.
\]
All other summands are nonnegative by Lemma~\ref{lem:cholesky}.
Summing over the first \(p-1\) passes and substituting the eigenvalue
formula for \(\beta_r\) gives
\[
\lambda^2E_n
\ge\frac q4\sum_{r=1}^{p-1}(\beta_r-\beta_{p-1})
=\frac14\sum_{l=0}^{q-1}\sum_{r=1}^{p-1}
\frac{(p-1-r)\omega_l^2}{(\lambda^2+r\omega_l)(\lambda^2+(p-1)\omega_l)}.
\]
For \(\omega_l\ge\lambda^2\), the denominator is at most
\(4r(p-1)\omega_l^2\). Keeping those terms, we obtain
\[
\lambda^2E_n
\ge\frac{1}{16}\sum_{l:\,\omega_l\ge\lambda^2}\sum_{r=1}^{p-1}
\rbr{\frac1r-\frac1{p-1}}
=\frac{H_{p-1}-1}{16}\,\abr{\cbr{l:\omega_l\ge\lambda^2}}.
\]

To count the eigenvalues above the threshold, we take the discrete
Fourier transform of \(\chi\) on the grid. Its absolutely convergent
Fourier series has coefficient \(\alpha_{|m|}/2\) at each frequency
\(m\ne0\), and
\(\sum_{j=0}^{q-1}e^{2\pi\mathrm{i}j(m-l)/q}\)
equals \(q\) when \(m\equiv l\pmod q\) and zero otherwise. Hence
\[
\omega_l
=\sum_{j=0}^{q-1}\chi(2\pi j/q)e^{-2\pi\mathrm{i}jl/q}
=\frac q2\sum_{\substack{m\ne0\\m\equiv l\pmod q}}\alpha_{|m|}.
\]
Keeping \(m=l\) for \(1\le l\le q-1\) gives
\(\omega_l\ge q/(2\zeta(2s)l^{2s})\). Thus every
\(l\in\{1,\ldots,q-1\}\) with
\(l\le(q/(2\zeta(2s)\lambda^2))^{1/(2s)}\) contributes to the count.
\end{proof}

\begin{lemma}\label{lem:info-gain}
Consider the kernel \eqref{eq:lifted-kernel} with \(s>1/2\).
For all \(n\ge1\) and \(\lambda>0\), let
\(l_n:=\ceil{\rbr{n/(\zeta(2s)\lambda^2)}^{1/(2s)}}\) and
\(C_s:=2(2s+\log 2+1/(2s-1))\). Then
\begin{align*}
\gamma_n
&\le\log\rbr{1+\frac{n}{\lambda^2}}+C_sl_n\\
&\le\log\rbr{1+\frac{n}{\lambda^2}}
+C_s\cbr{\rbr{\frac{n}{\lambda^2}}^{1/(2s)}+1}.
\end{align*}
For every deterministic sequence \(b_1,\ldots,b_n\in[0,2\pi]\), we also have
\[
\gamma_n\ge\log\det(I_n+\lambda^{-2}K_n^\circ)\ge\lambda^2E_n.
\]
\end{lemma}

\begin{proof}[Proof of Lemma~\ref{lem:info-gain}]
For arbitrary \(x_i=(a_i,b_i)\in\mathcal X\), write
\(a=(a_1,\ldots,a_n)^\top\), so \(K_n=K_n^\circ+aa^\top\).
Lemma~\ref{lem:rank-one-update} and
\((I_n+\lambda^{-2}K_n^\circ)^{-1}\preceq I_n\) give
\begin{align*}
\det(I_n+\lambda^{-2}K_n)
&=\det(I_n+\lambda^{-2}K_n^\circ)
\cbr{1+\lambda^{-2}a^\top(I_n+\lambda^{-2}K_n^\circ)^{-1}a}\\
&\le\det(I_n+\lambda^{-2}K_n^\circ)\rbr{1+\frac{n}{\lambda^2}},
\end{align*}
where we used \(\nbr{a}_2^2\le n\).

Let \(\Psi_m\in\mathbb R^{n\times 2m}\) have columns
\((\sqrt{\alpha_l}\cos(lb_i))_{i=1}^n\) and
\((\sqrt{\alpha_l}\sin(lb_i))_{i=1}^n\) for \(l\in[m]\).
Applying the two parts of Lemma~\ref{lem:determinant-tools}, we obtain
\[
\log\det(I_n+\lambda^{-2}\Psi_m\Psi_m^\top)
=\log\det(I_{2m}+\lambda^{-2}\Psi_m^\top\Psi_m)
\le 2\sum_{l=1}^m\log(1+n\alpha_l/\lambda^2).
\]
Here the two diagonal entries indexed by \(l\) are at most
\(1+n\alpha_l/\lambda^2\). The Fourier series converges absolutely, so
\(\Psi_m\Psi_m^\top\to K_n^\circ\) entrywise. Passing to the limit gives
\begin{equation}
\log\det(I_n+\lambda^{-2}K_n^\circ)
\le 2\sum_{l=1}^\infty\log(1+n\alpha_l/\lambda^2).
\label{eq:hadamard}
\end{equation}

Since \(n\alpha_l/\lambda^2\le(l_n/l)^{2s}\), we compare this series
with the integral of the decreasing function
\(t\mapsto\log(1+(l_n/t)^{2s})\) over \((0,\infty)\).
Rescaling by \(l_n\) gives
\begin{align*}
\sum_{l=1}^\infty\log(1+n\alpha_l/\lambda^2)
&\le l_n\int_0^\infty\log(1+u^{-2s})\,du\\
&\le l_n\rbr{\int_0^1(\log 2-2s\log u)\,du
+\int_1^\infty u^{-2s}\,du}\\
&=l_n\rbr{2s+\log 2+\frac{1}{2s-1}}.
\end{align*}
Combining \eqref{eq:hadamard} with the determinant estimate for the linear
coordinate and taking the supremum over \(x_1,\ldots,x_n\) gives the first
upper bound. The second follows from
\(l_n\le(n/\lambda^2)^{1/(2s)}+1\).

For the lower bound, take \(x_i=(0,b_i)\) so that \(K_n=K_n^\circ\).
By \eqref{eq:gamma}, the factorization \eqref{eq:GLD}, and
\eqref{eq:chol-diag}, we have
\begin{align*}
\gamma_n
&\ge\log\det(I_n+\lambda^{-2}K_n^\circ)
=\sum_{i=1}^n\log\frac{D_{ii}}{\lambda^2}\\
&=\sum_{i=1}^n\log\rbr{1+\frac{v_{<i}(b_i)}{\lambda^2}}
\ge\sum_{i=1}^n\frac{v_{<i}(b_i)}{\lambda^2+v_{<i}(b_i)}
\ge\lambda^2E_n.
\end{align*}
We used \(\log(1+u)\ge u/(1+u)\) for \(u\ge0\) and the representation
\eqref{eq:En-variance-form} with nonnegative subtracted terms.
\end{proof}

\subsection{Proof of the lower-bound theorem}

\begin{proof}[Proof of Theorem~\ref{thm:lower_bound}]
We first verify that the construction is an admissible instance.
The kernel sections at $(1,0)$ and $(-1,0)$ differ by the function
$(a,b)\mapsto2a$, and their squared distance in $\mathcal H_k$ is
$2+2-2\cdot0=4$. Thus,
\[
f_B=-\frac B2\{k(\cdot,(1,0))-k(\cdot,(-1,0))\},
\ \|f_B\|_{\mathcal H_k}=B.
\]
The Gaussian noise is $R$-sub-Gaussian, and the residual
$y_j+B\bar a_j=\epsilon_j$ is available after round $j$. Hence the rule
\eqref{eq:online-design} selects $x_i$ from past observations, and clipping
ensures $x_i\in\mathcal X$.

The scaling $\kappa$ is well defined for every $n\ge40$. Indeed, the
second pass repeats $b_1$, so Lemma~\ref{lem:cholesky} gives
\[
v_{-1}(b_1)\le v_{\{q+1\}}(b_1)
=1-\frac1{\lambda^2+1}<1=v_{<1}(b_1).
\]
The first summand of $E_n$ in \eqref{eq:En-variance-form} is therefore
positive, and all other summands are nonnegative.

We now lower bound $E_n$. Apply Lemma~\ref{lem:cyclic} with $p=20$ and
$q=\lfloor n/20\rfloor$. Since $H_{19}>7/2$,
\[
\lambda^2E_n
\ge\frac5{32}\min\{q-1,\lfloor T_q\rfloor\},
\ T_q:=\left(\frac{q}{2\zeta(2s)\lambda^2}\right)^{1/(2s)}.
\]
For fixed $s>1/2$ and $\lambda>0$, we have $T_q\to\infty$ and
$T_q/q\to0$. Choose an integer $n_1=n_1(s,\lambda)\ge\max\{40,\lambda^2\}$
such that $2\le T_q\le q$ for all $n\ge n_1$. Then
$\lfloor T_q\rfloor\ge T_q/2$ and $q-1\ge q/2\ge T_q/2$.
Using $q\ge n/40$, we obtain
\begin{equation}
E_n
\ge\frac5{64\lambda^2}
\left(\frac{n}{80\zeta(2s)\lambda^2}\right)^{1/(2s)}
=\frac{\bar c_s}{\lambda^2}
\left(\frac n{\lambda^2}\right)^{1/(2s)},
\ \bar c_s:=\frac5{64}\{80\zeta(2s)\}^{-1/(2s)}.
\label{eq:En-lower}
\end{equation}

The right-hand side grows polynomially in $n$, whereas
$\underline E(\lambda,1/11,n)$ in \eqref{eq:threshold} grows logarithmically.
We may therefore increase $n_1$ to an integer $n_2=n_2(s,\lambda)$ such
that $E_n\ge\underline E(\lambda,1/11,n)$ for every $n\ge n_2$.
Applying Lemma~\ref{lem:engine} with $\delta_0=1/11$, we obtain, for each
such $n$, with probability at least $6/11$,
\begin{equation}
\frac{|f_B(x_0)-\mu_n(x_0)|}{\sigma_n(x_0)}
\ge
\frac{4B}{21\sqrt3}+R\sqrt{\frac{E_n}{9408}}
\ge
c_s\left[B+\frac R\lambda
\left(\frac n{\lambda^2}\right)^{1/(4s)}\right],
\label{eq:first-inequality}
\end{equation}
where
$c_s:=\min\{4/(21\sqrt3),\sqrt{\bar c_s/9408}\}>0$.
The same event satisfies $\sigma_n^2(x_0)\ge2/21$.

To express this bound in terms of information gain, let $C_s$ be the
constant in Lemma~\ref{lem:info-gain}. There is an integer
$n_3=n_3(s,\lambda)$ such that, for all $n\ge n_3$,
$(n/\lambda^2)^{1/(2s)}\ge\max\{1,\log(1+n/\lambda^2)\}$.
The upper bound in that lemma then gives
\begin{equation}
\gamma_n\le(1+2C_s)\left(\frac n{\lambda^2}\right)^{1/(2s)}.
\label{eq:gamma-upper}
\end{equation}
Its lower bound and \eqref{eq:En-lower} show that, for
$n\ge n_0:=\max\{n_2,n_3\}$,
\begin{equation}
\bar c_s\left(\frac n{\lambda^2}\right)^{1/(2s)}
\le\gamma_n
\le(1+2C_s)\left(\frac n{\lambda^2}\right)^{1/(2s)}.
\label{eq:gamma-two-sided}
\end{equation}
In particular, $n_0\ge\lambda^2$, and both comparison constants depend
only on $s$. Set $c_s':=c_s/\sqrt{1+2C_s}$. Combining
\eqref{eq:first-inequality} and \eqref{eq:gamma-upper}, on the same event,
we have
\[
\frac{|f_B(x_0)-\mu_n(x_0)|}{\sigma_n(x_0)}
\ge c_s\left[B+\frac R\lambda
\left(\frac n{\lambda^2}\right)^{1/(4s)}\right]
\ge c_s'\left(B+\frac R\lambda\sqrt{\gamma_n}\right).
\]

Fix $\delta\in(0,1/2]$ and let $\rho_n(\delta)$ be a deterministic width
that gives coverage at least $1-\delta$ for this instance. If
$\rho_n(\delta)<c_s'(B+(R/\lambda)\sqrt{\gamma_n})$, then coverage fails
at $x_0$ throughout the event just constructed, whose probability is at
least $6/11>\delta$. This contradiction proves the claimed width bound.

For a data-dependent width with coverage at least $1-\delta$ on this
instance, the coverage event intersects our lower-bound event with
probability at least $6/11-\delta$. On that intersection the width is
at least $c_s'(B+(R/\lambda)\sqrt{\gamma_n})$, which gives the claim used
in Section~\ref{sec:discussion}.
\end{proof}

\section{Proof of the Minimax Characterization}
\label{app:minimax}

The dependence on the confidence level follows from an elementary Gaussian
tail bound.
\begin{lemma}
\label{lem:anticoncentration}
Let $g\sim\mathcal N(0,1)$. For every $\tau\ge0$,
\[
\PP(|g|>\tau)\ge\frac12e^{-\tau^2}.
\]
\end{lemma}
\begin{proof}[Proof of Lemma~\ref{lem:anticoncentration}]
Let $g'$ be an independent copy of $g$. Since $[-\tau,\tau]^2$ is contained
in the disk of radius $\sqrt2\tau$,
\[
\PP(|g|\le\tau)^2
=\PP(|g|\le\tau,\ |g'|\le\tau)
\le\PP(g^2+g'^2\le2\tau^2)
=1-e^{-\tau^2}.
\]
Here $g^2+g'^2$ has the exponential distribution with mean $2$. Taking the
complement and using $\sqrt{1-u}\le1-u/2$ for $u\in[0,1]$, we obtain
\[
\PP(|g|>\tau)\ge1-\sqrt{1-e^{-\tau^2}}\ge\frac12e^{-\tau^2}.
\]
\end{proof}

\begin{proof}[Proof of Corollary~\ref{cor:minimax}]
Fix $n\ge n_0(s,\lambda)$, $B,R>0$, and $\delta\in(0,1/4]$. Call a
deterministic width $\rho\ge0$ valid if the supremum of failure
probabilities in \eqref{eq:minimax-width} is at most $\delta$.
Theorem~\ref{thm:upper_bound} supplies such a width and gives
\[
\rho_n^*(\delta)
\le B+\frac R\lambda\sqrt{2\log\frac1\delta+\gamma_n}
\le\sqrt2\left(B+\frac R\lambda\sqrt{\gamma_n+\log\frac1\delta}\right).
\]

Every valid width must cover the hard instance of
Theorem~\ref{thm:lower_bound}. Since $\delta\le1/4<6/11$, its
constant-probability lower bound forces every such width to satisfy
$\rho\ge c_s'[B+(R/\lambda)\sqrt{\gamma_n}]$. Taking the infimum over
valid widths, we obtain
\[
\rho_n^*(\delta)\ge c_s'\left(B+\frac R\lambda\sqrt{\gamma_n}\right).
\]

To obtain the confidence-level term, consider the deterministic design
$x_i=x_\ddagger:=(0,0)$ for all $i\in[n]$, with $f\equiv0$ and i.i.d.\
noise $\epsilon_i\sim\mathcal N(0,R^2)$. This is an instance in
$\cP_n(B,R)$. Since $k(x_\ddagger,x_\ddagger)=1$, its kernel matrix and
test vector are $K_n=\one_n\one_n^\top$ and
$k_n(x_\ddagger)=\one_n$. The identity
$(\lambda^2I_n+K_n)\one_n=(\lambda^2+n)\one_n$ and
\eqref{eq:posterior} yield
\[
\mu_n(x_\ddagger)=\frac{\sum_{i=1}^n\epsilon_i}{\lambda^2+n},
\ 
\sigma_n^2(x_\ddagger)=\frac{\lambda^2}{\lambda^2+n}.
\]
Consequently,
\[
\frac{f(x_\ddagger)-\mu_n(x_\ddagger)}{\sigma_n(x_\ddagger)}
=-\frac{\sum_{i=1}^n\epsilon_i}{\lambda\sqrt{\lambda^2+n}}
\sim\mathcal N(0,\varsigma_n^2),
\ 
\varsigma_n^2:=\frac{R^2n}{\lambda^2(\lambda^2+n)}
\ge\frac{R^2}{2\lambda^2},
\]
where $n\ge n_0\ge\lambda^2$. A valid width has failure probability at most
$\delta$ at $x_\ddagger$. Applying Lemma~\ref{lem:anticoncentration}, we
therefore find
\[
\delta\ge\PP\left(|g|>\frac{\rho}{\varsigma_n}\right)
\ge\frac12\exp\left(-\frac{\rho^2}{\varsigma_n^2}\right),
\ g\sim\mathcal N(0,1).
\]
Since $\delta\le1/4$, we have
$\log(1/(2\delta))\ge\frac12\log(1/\delta)$, and hence
\[
\rho^2\ge\varsigma_n^2\log\frac1{2\delta}
\ge\frac{R^2}{4\lambda^2}\log\frac1\delta.
\]
This holds for every valid $\rho$, so
$\rho_n^*(\delta)\ge(R/(2\lambda))\sqrt{\log(1/\delta)}$.

Set $\underline c_s:=\frac12\min\{c_s',1/2\}$. By taking the average of
the two lower bounds, we obtain
\begin{align*}
\rho_n^*(\delta)
&\ge\frac12\left[
 c_s'\left(B+\frac R\lambda\sqrt{\gamma_n}\right)
 +\frac R{2\lambda}\sqrt{\log\frac1\delta}\right]\\
&\ge\underline c_s\left[B+\frac R\lambda
 \left(\sqrt{\gamma_n}+\sqrt{\log\frac1\delta}\right)\right]\\
&\ge\underline c_s\left(B+\frac R\lambda\sqrt{\gamma_n+\log\frac1\delta}\right).
\end{align*}
Together with the upper bound, this proves the claimed comparison, with
constants depending only on $s$.
\end{proof}

\end{document}